\documentclass[times,twocolumn,final]{elsarticle}

\usepackage{medima}
\usepackage{framed,multirow}
\usepackage{booktabs}
\usepackage{amsmath,amssymb,amsfonts}
\usepackage{algorithmic}
\usepackage{textcomp}
\usepackage{latexsym}
\usepackage[ruled,linesnumbered,noline]{algorithm2e}

\usepackage{url}
\usepackage{xcolor}
\usepackage{subfig}
\usepackage{float}
\usepackage{hyperref}
\hypersetup{hypertex=true,
colorlinks = true,
linkcolor = blue,
citecolor = blue}
\usepackage{graphicx}

\definecolor{newcolor}{rgb}{.8,.349,.1}

\journal{Medical Image Analysis}

\begin{document}

\verso{Wei Li \textit{et~al.}}

\begin{frontmatter}

\title{F\textsuperscript{2}TTA: Free-Form Test-Time Adaptation on Cross-Domain Medical Image Classification via Image-Level Disentangled Prompt Tuning}

\author[1,2,4]{Wei Li}

\author[3]{Jingyang Zhang \corref{cor1}}
\ead{zjysjtu1994@gmail.com}

\author[4]{Lihao Liu}
\author[5]{Guoan Wang}
\author[4]{Junjun He}
\author[3]{Yang Chen}

\author[1,2]{Lixu Gu \corref{cor1}}
\ead{gulixu@sjtu.edu.cn}

\cortext[cor1]{Corresponding author.}
  


\address[1]{School of Biomedical Engineering, Shanghai Jiao Tong University, Shanghai, 200240, China}
\address[2]{Institute of Medical Robotics, Shanghai Jiao Tong University, Shanghai, 200240, China}
\address[3]{School of Computer Science and Engineering, Southeast University, Nanjing, 210096, China}
\address[4]{Shanghai Artificial Intelligence Laboratory, Shanghai, 200000, China}
\address[5]{Department of Systems and Enterprises, Stevens Institute of Technology, Hoboken, NJ, 07030, United States}


\begin{abstract}
Test-Time Adaptation (TTA) has emerged as a promising solution for adapting a source model to unseen medical sites using unlabeled test data, due to the high cost of data annotation.
Existing TTA methods consider scenarios where data from one or multiple domains arrives in complete domain units.
However, in clinical practice, data usually arrives in domain fragments of arbitrary lengths and in random arrival orders, due to resource constraints and patient variability.
This paper investigates a practical Free-Form Test-Time Adaptation (F$^{2}$TTA) task, where a source model is adapted to such free-form domain fragments, with shifts occurring between fragments unpredictably.
In this setting, these shifts could distort the adaptation process.
To address this problem, we propose a novel Image-level Disentangled Prompt Tuning (I-DiPT) framework. 
I-DiPT employs an image-invariant prompt to explore domain-invariant representations for mitigating the unpredictable shifts, and an image-specific prompt to adapt the source model to each test image from the incoming fragments.
The prompts may suffer from insufficient knowledge representation since only one image is available for training.
To overcome this limitation, we first introduce Uncertainty-oriented Masking (UoM), which encourages the prompts to extract sufficient information from the incoming image via masked consistency learning driven by the uncertainty of the source model representations.
Then, we further propose a Parallel Graph Distillation (PGD) method that reuses knowledge from historical image-specific and image-invariant prompts through parallel graph networks.
Experiments on breast cancer and glaucoma classification demonstrate the superiority of our method over existing TTA approaches in F$^{2}$TTA.
Code is available at \url{https://github.com/mar-cry/F2TTA}.
\end{abstract}

\begin{keyword}
\KWD Test-Time Adaptation\sep Prompt Tuning \sep Masked Image Modeling \sep Graph Neural Networks
\end{keyword}

\end{frontmatter}



\section{Introduction}
\label{sec:intro}

Recent deep learning-based medical image classification methods
have achieved superior performance, demonstrating their value in disease
diagnosis \cite{pathologydisease,medicalclass,transformer4CLS}. 
Their success relies on an assumption that the training and test data share similar distributions.
However, in healthcare, the classification model is often required to be applied
to multiple medical sites or hospitals (i.e., domains) that differ from the training domains \cite{multi-site}. 
Therefore, domain shifts may occur between the training datasets (i.e., source domains) and test datasets (i.e., target domains), primarily due to the diversity of imaging devices, protocols, and patient groups from various domains \cite{shiftscause}. 
Such shifts can substantially degrade the model's performance at test time \cite{DAreview}. 
Test-time adaptation (TTA) \cite{tent} has emerged as a promising paradigm to mitigate such domain shifts by adapting the model to target domains using unlabeled test data, as collecting and annotating sufficient target domain data in advance is time-consuming and labor-intensive \cite{cscada}.

\begin{figure}[!t]
    \centerline{\includegraphics[width=\columnwidth]{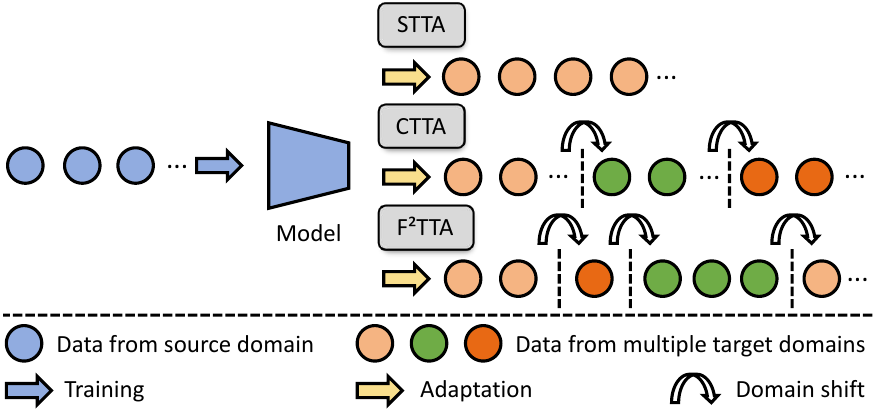}}
    \caption{Comparison of different TTA schemes. STTA addresses adaptation on a
    single target domain under a static domain shift, whereas CTTA assumes
    periodic shifts between complete domains during adaptation across multiple
    target domains. In contrast, F$^{2}$TTA considers a practical
    scenario where data arrives in random domain fragments with unpredictable
    shifts between the fragments.}
    \label{fig:setting}
\end{figure}

A natural insight is to leverage Single-domain TTA (STTA) methods \cite{tent,dltta,t3a,ttt} to adapt a model trained on the source domain (i.e., source model) to each target domain.
STTA assumes that there exist only static domain shifts between the source domain and each target domain.
When applied to continuously changing target domains, these methods inevitably suffer from error accumulation due to domain shifts between target domains \cite{cotta}. 
In contrast, Continual TTA (CTTA) \cite{cotta,ctta-mia,ctta-cvpr24,cmae} is proposed to address the limitation of STTA on continuously changing target domains by learning target domain representations while maintaining robustness against the shifts between target domains.
Despite its success, CTTA considers only a simplified scenario, where data arrives sequentially in complete domains, with shifts occurring periodically between these complete domains. 
However, in clinical practice, traversing all data within each domain at once is highly impractical due to resource limitation \cite{datalimit} and patient variability \cite{datalimit2}. 
Alternatively, it is feasible to prioritize collecting data fragments rather than the complete datasets from different domains.
In this way, data tends to arrive in random domain fragments without predefined fragment lengths or arrival orders, leading to unpredictable domain shifts across these fragments, as shown in Fig. \ref{fig:setting}.
In this scenario, such unpredictable shifts would disrupt the representation learning of existing CTTA methods that rely heavily on complete domains with only mild shifts, hindering stable adaptation especially when dealing with continuously arriving fragments.

To this end, we study a practical \emph{Free-Form Test-Time Adaptation (F$^{2}$TTA)} paradigm that focuses on adaptation to domain fragments of arbitrary lengths and random arrival orders, without specific constraints on the domain form.
The key challenge of F$^{2}$TTA is overcoming the unpredictable shifts occurring between free-form fragments. 
To enhance robustness against such shifts, it is crucial to learn domain-invariant representations \cite{cddsa,disentangle-1,disentangle-2} from continuously arriving fragments.
Following this insight, prior work has proposed domain-level disentangled prompt tuning \cite{pt4cl1,prompttta,prompt-graph} for CTTA, where a domain-invariant prompt is maintained across all domains to capture the invariant representations and a domain-specific prompt is applied to adapt the source model to each target domain. 
However, it is impractical to apply domain-level prompts in F$^{2}$TTA, as they rely on complete domain data to learn stable representations, and could fail when encountering unpredictable and random fragments \cite{dpcore}.
To address this problem, a promising way is to utilize an image-specific prompt to adapt the model to each image and retain an image-invariant prompt to learn the invariant representations.
Despite its potential, leveraging the image-level prompts often suffers from insufficient knowledge representation, especially when only one image is available for training the prompts \cite{ctta-cvpr24}.
We propose to mitigate the above limitation in two aspects:
1) extracting potential information from a single incoming image; 
2) reusing historical knowledge contained in previous prompts.

For the first aspect, the spatial context relations can serve as important additional clues to enhance the perception capability of the image-level disentangled prompts \cite{mim-random1}.
This motivation is inspired by the fact that pathologists typically leverage both the local information around a single area and the global correlations across different areas to make diagnoses in whole slide image analysis \cite{pathexample}.
To achieve this, we seek to adopt Masked Image Modeling (MIM) \cite{mae}, where a subset of image patches is masked to encourage both prompts to explore knowledge contained in the masked regions based on contextual information from the unmasked patches.
Considering the substantial semantic difference between the image-specific and image-invariant prompts, we need to select appropriate regions to mask for each type of prompt.
Recent studies indicate that the invariant representations are stable under domain shifts \cite{d2ct}, whereas the image-specific representations tend to be sensitive to the shifts \cite{UDA}.
Motivated by the above observation, we propose to utilize the representation uncertainty of image patches to guide the masking process, thereby encouraging the image-invariant and image-specific prompts to extract the respective knowledge from the corresponding masked regions.

For the second aspect, historical prompts may encode image patterns similar to the incoming image, providing prior knowledge for the current prompts to improve adaptation performance \cite{param_init}.
Graph Neural Networks (GNNs) \cite{velivckovic2017graph} emerge as a feasible approach to understand the intricate relationships among the prompts and distill useful components to facilitate the representation process of the current prompts.
In the light of the aforementioned semantic differences between the two types of prompts, we design parallel customized graphs to extract distinct knowledge independently.

In this paper, we propose \emph{to our knowledge the first F$^{2}$TTA work} for cross-domain medical image classification, where images from different domains arrive in free-form fragments of arbitrary lengths and random arrival orders. 
Our framework is called Image-level Disentangled Prompt Tuning (I-DiPT), aiming to adapt a source model to each target image through an image-specific prompt, while overcoming the unpredictable shifts by using an image-invariant prompt to explore domain-invariant representations.
To mitigate the insufficient representation when only a single image is available to update the prompts, we first propose Uncertainty-oriented Masking (UoM), which encourages the image-invariant and image-specific prompts to fully extract their respective knowledge from the incoming image. 
Specifically, we mask image patches based on the representation uncertainty of image features and enforce the prompts to instruct the model to make consistent predictions between the unmasked patches and the full image.
Then, we propose a Parallel Graph Distillation (PGD) mechanism to further enhance the representation by reusing knowledge from historical prompts.
PGD constructs two prompt graphs to distill knowledge from previous image-specific and image-invariant prompts, respectively. The distilled knowledge is then injected into the prompts for the incoming image.

We summarize the contributions as follows:
\begin{itemize}
    \item We study a novel and practical Free-Form Test-Time Adaptation (F$^{2}$TTA)
        task and propose an Image-level Disentangled Prompt Tuning (I-DiPT) framework to adapt a classification model to free-form fragments from multiple domains under
        unpredictable domain shifts.

    \item We propose an Uncertainty-oriented Masking (UoM) scheme to encourage the image-level prompts to harvest sufficient information from the incoming image.

    \item We develop a Parallel Graph Distillation (PGD) method to retain historical knowledge contained in previous prompts for effective adaptation.

    \item We conduct extensive experiments on the breast cancer histology image and
        the glaucoma fundus image classification, demonstrating the superiority
        of our method over state-of-the-art STTA and CTTA methods in F$^{2}$TTA.
\end{itemize}

\section{Related Work}
\label{sec:Related Work}
\subsection{Single-domain Test-Time Adaptation (STTA)}
Test-Time Adaptation (TTA) aims to adapt a model to test datasets using only unlabeled test data, as distribution shifts often occur between training and test sets. 
These shifts typically arise because test datasets are collected from various sites using different imaging devices or protocols \cite{cddsa}. 
Single-domain TTA (STTA) \cite{tent} focuses on adapting the model to a single target domain under a static domain shift. 
Existing STTA methods mainly leverage self-supervised learning schemes \cite{tent,dltta,t3a,ttt}. 
For example, Tent \cite{tent} utilizes entropy minimization on model predictions as the adaptation loss and updates batch normalization layers to reduce computational overhead. 
T3A \cite{t3a} performs classifier adjustment through class prototype representation learning. 
DLTTA \cite{dltta} further accounts for image-specific distribution shifts and dynamically adjusts the learning rate for each test sample. 
However, error accumulation \cite{cotta} inevitably occurs when directly applying STTA methods to scenarios with multiple target domains, as they overlook the distribution shifts between target domains.

\subsection{Continual Test-Time Adaptation (CTTA)}
Continual Test-Time Adaptation (CTTA) has been proposed to enable model adaptation across multiple target domains. 
Existing CTTA methods mainly focus on enhancing the robustness of supervision \cite{cotta,ctta-mia,sar} or mitigating domain style discrepancies \cite{ctta-cvpr24,ctta-featurealign,cmae,dpcore} to address the limitations of Single-step TTA (STTA). 
For example, CoTTA \cite{cotta} leverages weight- and augmentation-averaged pseudo labels to learn target domain knowledge and randomly resets a portion of model parameters to their source-initialized values to prevent knowledge overwriting. 
SAR \cite{sar} excludes high-entropy test samples that could cause large gradient spikes, then applies sharpness-aware entropy minimization on the remaining data to steer the model weights toward a flatter minimum. 
VPTTA \cite{ctta-cvpr24} introduces image-specific visual prompts and employs a feature alignment loss to bridge the gaps across domains. 
BECoTTA \cite{becotta} assumes that the domain shifts occur gradually over time (i.e., data from two adjacent domains may arrive concurrently) and proposes a mixture-of-domain low-rank expert model to simultaneously handle data from both domains. 
DPCore \cite{dpcore} further considers dynamic domain changes caused by the unpredictability of weather conditions and adaptively adjusts visual prompts based on domain similarity between each incoming target batch and stored source domain samples. 
Despite their success, these CTTA methods assume that data arrives sequentially in complete domain units, which does not hold in our F$^{2}$TTA setting due to unpredictable shifts introduced by data arriving in free-form domain fragments, as discussed in Sec. \ref{sec:intro}.

\subsection{Prompt Tuning (PT)}
Prompt tuning was first introduced to provide additional textual instructions for pre-trained models to perform downstream tasks in natural language processing (NLP) \cite{prompttuning}. 
Inspired by its success in NLP, recent studies \cite{pt4tta1,ctta-cvpr24,pt4cl1,prompttta,pt4medicalseg,svdp} have attempted to transfer prompt tuning to computer vision tasks by introducing and tuning a small number of parameters while keeping the backbone model frozen. 
For example, VPTTA \cite{ctta-cvpr24} modulates the style of each input image in the frequency domain using image-specific visual prompts to mitigate domain shifts. 
SVDP \cite{svdp} proposes adaptively allocating learnable parameters on the pixels and designs an image-specific prompt optimization strategy. 
However, prompts designed for specific images often fail to capture features shared across different domains. 
To address this, DualPrompt \cite{pt4cl1} introduces domain-invariant and domain-specific prompts into the model to mitigate catastrophic forgetting caused by domain shifts and to enhance generalization. 
Gan et al. \cite{prompttta} employ the domain-level disentangled prompt tuning in the CTTA setting to learn domain-invariant representations during adaptation. However, in F$^{2}$TTA, the domain identity of each test sample is difficult to acquire in advance, limiting the applicability of domain-level prompts. 
In contrast, the proposed I-DiPT introduces image-level disentangled prompts to learn invariant embeddings, enabling effective adaptation to free-form domain fragments and mitigating unpredictable domain shifts.

\section{Methodology}
\label{sec:method}

\begin{figure*}[!htbp]
\centerline{\includegraphics[width=\textwidth]{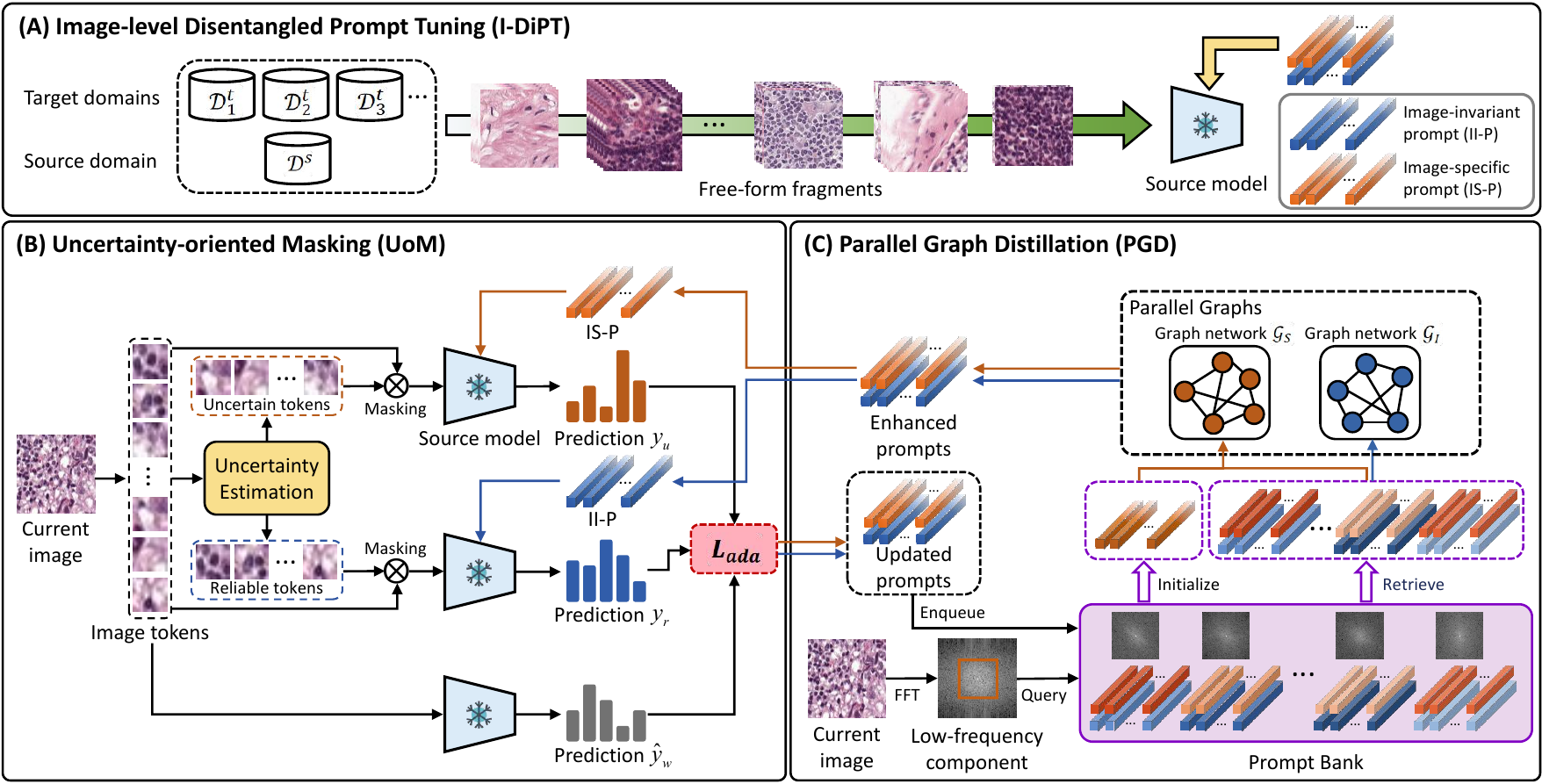}}
\caption{Overview of the proposed Image-level Disentangled Prompt Tuning (I-DiPT) framework for Free-Form Test-Time Adaptation (F$^{2}$TTA).
For an incoming image, we adapt the source model to this image from the incoming domain fragments using an image-specific prompt, and learn invariant representations guided by an image-invariant prompt to mitigate the unpredictable shifts in F$^{2}$TTA (Section~\ref{sec:idipt}).
We address the problem of insufficient representation in image-level prompts in two ways:  
1) We propose Uncertainty-oriented Masking (UoM) to encourage the prompts to extract sufficient information from the incoming image via masked consistency learning (Section~\ref{sec:uom}); and  
2) We propose Parallel Graph Distillation (PGD) to retain knowledge from historical prompts, facilitating the learning of prompts for the incoming image (Section~\ref{sec:pgd}).
}
\label{fig:main_figure}
\end{figure*}

In the Free-Form Test-Time Adaptation (F\(^2\)TTA), a model trained on a source domain $\mathcal{D}^s$ is required to adapt to unlabeled test data arriving in domain fragments $\{\mathcal{R}_{ij}\}_{i\in [1,N], j\in [1,M_i]}$, where $\mathcal{R}_{ij}$ denotes the $j^{th}$ fragment of the $i^{th}$ target domain $\mathcal{D}^t_i$, and $M_i$ is the number of fragments in domain $\mathcal{D}^t_i$. 
The fragments arrive in a random order, and the length of each fragment $\mathcal{R}_{ij}$, denoted as $N^f_{ij}$, is random. These lengths satisfy the constraint $\sum_{i=1}^{N} \sum_{j=1}^{M_i} N^f_{ij} = N_T$, where $N_T$ denotes the total number of test images.
Notably, fragments from the source domain also appear during the adaptation process, as the model needs to process images from the source domain in clinical practice. 
Our goal is to adapt the source model to these free-form fragments while maintaining its performance on the source domain under domain shifts between the fragments.

In our framework, although data arrives in fragments, adaptation is performed for each test image $x_{k,k\in[1,N_T]}$ from the incoming fragment based on image-level prompts.
Fig. \ref{fig:main_figure} demonstrates the overall framework for F\(^2\)TTA.  
Specifically, we propose Image-level Disentangled Prompt Tuning (I-DiPT), where we leverage an image-specific prompt to adapt the source model to each test image and maintain an image-invariant prompt to learn domain-invariant representations, thereby mitigating the unpredictable shifts between fragments.
To mitigate the limited data available for updating the prompts, we first propose Uncertainty-oriented Masking (UoM) to encourage the image-level disentangled prompts to effectively extract the corresponding knowledge from the incoming image.
Then, we propose a Parallel Graph Distillation (PGD) mechanism to distill knowledge from historical prompts to enhance the prompts for the incoming image.

\subsection{Image-level Disentangled Prompt Tuning (I-DiPT)}
\label{sec:idipt}
In the F\(^2\)TTA setting, test images arrive in free-form domain fragments with unpredictable domain shifts.
Existing disentangled prompt tuning methods \cite{pt4cl1,prompttta,prompt-graph} employ domain-level disentangled prompts to adapt the source model to each target domain while overcoming the domain shifts.
However, the domain identity of incoming fragments is difficult to acquire in advance, making domain-level prompts impractical in F\(^2\)TTA.
Moreover, the domain-level prompts rely on the complete domain data to learn stable representations, leading to noisy adaptation when encountering domain fragments.
Therefore, we propose Image-level Disentangled Prompt Tuning (I-DiPT) to mitigate the unpredictable shifts during the adaptation.

\subsubsection{Prompt Setup} 
We construct a vision transformer (ViT) backbone \cite{vit} as a classification model $f_{\theta}(\cdot)$. In ViT, an input image is converted into a sequence of flattened patches (or tokens) and a classification prediction is obtained by mapping these tokens through transformer layers. 
During adaptation, for the current image $x_k$, we define image-specific and image-invariant prompts as $(\phi^{(k)}_S \in \mathbb{R}^{L_S \times C},\ \phi^{(k)}_I \in \mathbb{R}^{L_I \times C})$, where $L_S$ and $L_I$ denote sequence lengths of the two prompts, respectively, and $C$ is the feature dimension of the ViT.
Notably, the notation $\phi^{(k)}_{I}$ denotes the image-invariant prompt learned during the adaptation for the current image $x_k$, which shares parameters across all test images.
We insert the prompts into the multi-head self-attention (MSA) layers of the transformer blocks. 
We select the prefix tuning strategy \cite{pt4cl1} because it does not alter the length of the input and output sequences, allowing the prompts to be inserted into multiple MSA layers in a scalable and efficient way.

\begin{figure}[!htbp]
\centerline{\includegraphics[width=\linewidth]{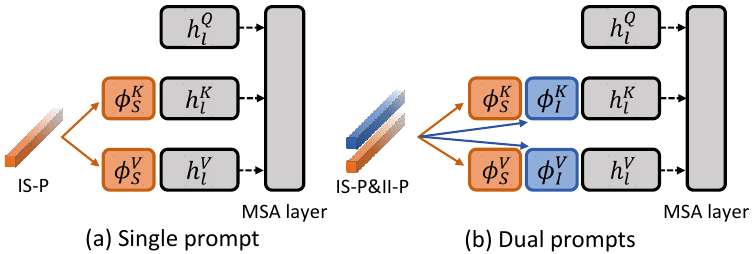}}
\caption{Visualization of prompt embedding into the MSA layer. (a) embedding of a single prompt (e.g., image-specific prompt, IS-P); (b) joint embedding of IS-P and II-P (image-invariant prompt).}
\label{fig:prompt_setup}
\end{figure}

Specifically, we take the image-specific prompt $\phi^{(k)}_S$ as an example and abbreviate it as $\phi_S$ for brevity.
For the $l^{th}$ MSA layer, the prompt $\phi_S$ is first split into two sub-prompts $\phi^{K}_S, \phi^{V}_S \in \mathbb{R}^{L_S / 2 \times C}$, and then prepended to the key and value of this MSA layer, respectively:
\begin{equation}
    h^{o}_l = f^l_{MSA} (h^{Q}_l, [\phi^{K}_S; h^{K}_l], [\phi^{V}_S; h^{V}_l]),
\end{equation}
where $f^l_{MSA}$ represents the $l^{th}$ MSA layer, $h^{Q}_l, h^{K}_l, h^{V}_l, h^{o}_l$ denote the input query, key, value, and output of this MSA layer, and $[;]$ denotes the concatenation operation along the sequence length dimension. 
The image-invariant prompt $\phi_I$ can also be attached to the MSA layer in the same way.
When the two prompts are attached to the same MSA layer simultaneously, the sub-prompts from each prompt should first be concatenated before being attached to this layer:
\begin{equation}
    h^{o}_l = f^l_{MSA} (h^{Q}_l, [[\phi^{K}_S;\phi^{K}_I]; h^{K}_l], [[\phi^{V}_S;\phi^{V}_I]; h^{V}_l]),
\end{equation}
where $\phi^{K}_I, \phi^{V}_I$ denote the sub-prompts of $\phi^{(k)}_{I}$. 
Fig. \ref{fig:prompt_setup} illustrates how the prompts are embedded into the MSA layer.

\subsubsection{Prompt Tuning} 
We attach both prompts to the model $f_\theta$ and utilize an adaptation loss $\mathcal{L}_{ada}$ to update the prompts:
\begin{equation}
    \label{eq:overall_loss}
    \mathop{\mathrm{min}}_{\phi^{(k)}_I,\phi^{(k)}_S} \mathcal{L}_{ada}(f_\theta,x_k,\phi^{(k)}_I,\phi^{(k)}_S).
\end{equation}
After updating the prompts, both the image-invariant and image-specific prompts are inserted into the model, denoted as $f_{\phi^{(k)}_I, \phi^{(k)}_S}$, to make predictions $\hat{y}_a$ for the current image $x_k$.

\subsection{Uncertainty-oriented Masking (UoM)}
\label{sec:uom}
During adaptation, the adaptation loss is crucial for driving the prompts to extract necessary information from the current image. 
Existing TTA and CTTA methods mainly rely on entropy minimization \cite{tent} or pseudo-labeling \cite{cotta} to drive parameter updates.
However, the image-level prompts with limited learnable parameters may struggle to acquire sufficient knowledge without ground-truth guidance when driven by the above loss functions.
Therefore, we propose Uncertainty-oriented Masking (UoM) to encourage the image-level prompts to effectively extract potential information from the incoming image for adaptation based on the spatial context relationships.
We mask specific image tokens based on the representation uncertainty and encourage the prompts to instruct the model to make predictions consistent with those obtained from the entire image, using only the unmasked tokens.

\subsubsection{Uncertainty Estimation Mechanism}
\begin{figure}[!htbp]
\centerline{\includegraphics[width=\linewidth]{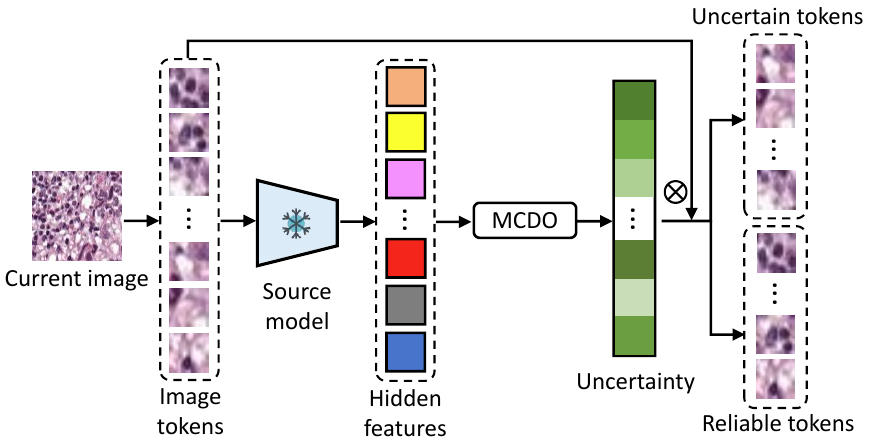}}
\caption{The process of the uncertainty estimation mechanism. The $\otimes$ indicates that uncertain and reliable tokens are selected from image tokens based on feature uncertainty.}
\label{fig:uom}
\end{figure}

We utilize Monte Carlo DropOut (MCDO) \cite{uncertainty} to quantify the token-wise uncertainty of the source model representations. 
Specifically, given an input image $x_k$, we perform $D$ forward passes through the source model to obtain a set of token-wise hidden features $\{E_d \in \mathbb{R}^{L \times C} \}^D_{d=1}$ under random dropout, where $L$ denotes the number of image tokens. 
Then we estimate the uncertainty $U_j$ of the $j^{th}$ token $(j\in[1:L])$ based on the obtained features:
\begin{equation}
U_j = \Bigg ( \frac{1}{D} \sum^D_{d=1} \Vert p(e_{dj}) - \mu_j  \Vert^2 \Bigg )^{\frac{1}{2}},
\end{equation}
where $e_{dj} \in \mathbb{R}^{1 \times C}$ denotes the embeddings of the $j^{th}$ token extracted from the hidden feature $E_d$ in the $d^{th}$ forward pass, $p(\cdot)$ denotes the average pooling operation that converts the feature $e_{dj}$ to a scalar value, and $\mu_j$ is the average over $D$ forward passes for the $j^{th}$ token, i.e., $\mu_j = \frac{1}{D} \sum^D_{d=1} p(e_{dj})$. 
Notably, MCDO is applied only to the feedforward network in the first transformer layer for computational efficiency \cite{cmae}. 
Based on the uncertainty $U_j$, we select the top $K\%$ tokens with the highest uncertainty as uncertain tokens $\{x^{(j)}_k\}_{j\in \mathcal{T}_{U}}$, while the top $K\%$ tokens with the lowest uncertainty are considered as reliable tokens $\{x^{(j)}_k\}_{j\in \mathcal{T}_{R}}$, where $x^{(j)}_k$ denotes the $j^{th}$ flattened image token, and $\mathcal{T}_{U}$, $\mathcal{T}_{R}$ denote the sets of uncertain and reliable token IDs, respectively.
The above process is demonstrated in Fig. \ref{fig:uom}.

\subsubsection{Masked Consistency Learning} 
\label{sec:mased_learning}
After masking the uncertain tokens, the remaining tokens $\{x^{(j)}_k\}_{j\notin \mathcal{T}_{U}}$ are then fed into the model attached with an image-specific prompt $\phi^{(k)}_S$ to obtain the classification prediction $y_u = f_{\phi^{(k)}_S}(\{x^{(j)}_k\}_{j\notin \mathcal{T}_{U}})$. 
A consistency constraint is then imposed between $y_u$ and the prediction obtained by feeding the entire image into the source model $\hat{y}_w = f_{\theta}(\{x^{(j)}_k\}_{j\in[1:L]})$.
Meanwhile, to update the image-invariant prompt, the same consistency learning strategy is also applied between the predictions from reserved tokens after masking reliable tokens $y_r = f_{\phi^{(k)}_I}(\{x^{(j)}_k\}_{j\notin \mathcal{T}_{R}})$ and the predictions from the entire image. 
We utilize cross-entropy loss $\mathcal{L}_{ce}$ as the consistency constraint, and the overall learning objective is as follows:
\begin{equation}
    \mathcal{L}_{ada} = \mathcal{L}_{ce}(y_u, \hat{y}_w) + \mathcal{L}_{ce}(y_r, \hat{y}_w).
    \label{eq:uom}
\end{equation}

\subsection{Parallel Graph Distillation (PGD)}
\label{sec:pgd}
In DiPT, the image-specific prompt is re-initialized for the current image and the image-invariant prompt would encounter potential knowledge forgetting due to appearance heterogeneity of test images \cite{forget}.
The historical prompts encode similar image patterns that can serve as prior knowledge to benefit the adaptation to the current image.
Therefore, we propose a Parallel Graph Distillation (PGD) method to distill and retain knowledge from historical prompts to improve the adaptation performance for the current image.

\subsubsection{Prompt Bank} 
We first construct a prompt bank to store the historical prompts.
Specifically, the prompt bank $\mathcal{B}$ stores $N_B$ keys and their corresponding values, i.e., $\mathcal{B}=\{(k_b, v_b)\}^{N_B}_{b=1}$. 
The key $k_b$ is the low-frequency component within the amplitude spectrum of a previous image $x \in \mathbb{R}^{H\times W\times 3}$ in the adaptation process. 
The value $v_b=(\phi^{(b)}_S, \phi^{(b)}_I)$ represents the stored prompts.
The low-frequency component $\mathcal{A}_{L} (x)$ of image $x$ can be extracted via Fast Fourier Transform (FFT) \cite{fft}, i.e., $\mathcal{A}_{L} (x) = \mathcal{A}(x)\cdot \mathcal{M}$, where $\mathcal{A}(x)$ represents the amplitude spectrum of $x$ and $\mathcal{M} = \mathbb{I}_{(h,w)\in[-\beta H:\beta H, -\beta W:\beta W]}$ is a binary mask to control the scale of the low-frequency component at the $\beta$ level. 
The value of $\mathcal{M}$ is 1 at the central region and 0 elsewhere.
The prompt bank holds the First In First Out (FIFO) principle. 

\subsubsection{Parallel Graph Distillation}
After constructing the prompt bank, we construct two graph networks $\mathcal{G}_S$, $\mathcal{G}_I$ for image-specific and image-invariant prompts separately due to the semantic differences between the two prompts. 
Taking the distillation process for the image-specific prompt as an example, we first extract the low-frequency component of the current image $\mathcal{A}_{L}(x_k)$ and then compute the cosine similarity $\omega_b$ between $\mathcal{A}_{L}(x_k)$ and each key $k_b$ in the prompt bank. 
Subsequently, we pre-initialize an image-specific prompt $\tilde{\phi}^{(k)}_S$ for the current image via $\tilde{\phi}^{(k)}_S = \sum^{N_B}_{b=1} \omega_b v_b$ to accelerate convergence. 
The pre-initialization process actually encodes knowledge from historical prompts. 
However, such a simple linear combination cannot capture the complex semantic correlations among the prompts. 
We further distill knowledge from historical prompts using a graph network $\mathcal{G}_S$.
The pre-initialized prompt $\tilde{\phi}^{(k)}_S$ and the stored prompts $\{ \phi^{(b)}_S \}_{b=1}^{N_B}$ are transformed into graph nodes $n^{(k)}_{S} = W^{e}_S \tilde{\phi}^{(k)}_S$ and $n^{(b)}_{S} = W^{e}_S \phi^{(b)}_S$ through a learnable linear transformation $W^{e}_S: \mathbb{R}^{L_S \times C} \rightarrow \mathbb{R}^{D^g_S}$, where $D^g_S$ denotes the dimension of the graph node embeddings.
Then, we use a fully connected layer $V_S:\mathbb{R}^{D^g_S} \times \mathbb{R}^{D^g_S} \to \mathbb{R}^{1}$ to compute an attention coefficient $a^{(kb)}_{S}$ between nodes $n^{(k)}_S$ and $n^{(b)}_S$:
\begin{equation}
    \label{eq:attention_coeff}
   a^{(kb)}_{S} = \frac{\exp (V_S (n^{(k)}_S, n^{(b)}_S))}{\sum^{N_B}_{b=1} \exp (V_S (n^{(k)}_S, n^{(b)}_S))}, 
\end{equation}
The enhanced prompt $\hat{\phi}^{(k)}_S$ is obtained as follows: 
\begin{equation}
\label{eq:specific_graph_prompt}
  \hat{\phi}^{(k)}_S = \tilde{\phi}^{(k)}_S + W^d_S \Big ( \sum^{N_B}_{b=1} a^{(kb)}_S n^{(b)}_S \Big ),  
\end{equation}
where $W^d_S: \mathbb{R}^{D^g_S} \to \mathbb{R}^{L_S \times C}$ is a learnable linear transformation used to reconstruct the graph nodes to the prompts.
We utilize another graph network $\mathcal{G}_I$ to distill knowledge from previous image-invariant prompts and introduce it into the current image-invariant prompt $\hat{\phi}^{(k)}_I$ through the Exponential Moving Average strategy to avoid knowledge overwriting:
\begin{equation}
\label{eq:invariant_graph_prompt}
\begin{aligned}
    a^{(kb)}_{I} &= \frac{\exp (V_I (n^{(k)}_I, n^{(b)}_I))}{\sum^{N_B}_{b=1} \exp (V_I (n^{(k)}_I, n^{(b)}_I))},   \ \text{and} \\ 
    \hat{\phi}^{(k)}_I &= \gamma \phi^{(k)}_I + (1-\gamma) W^d_{I} \Big( \sum^{N_B}_{b=1} a^{(kb)}_I n^{(b)}_I \Big),
\end{aligned}    
\end{equation}
where
$a^{(kb)}_I$ is the attention coefficient between node pairs;
$V_I: \mathbb{R}^{D^g_I} \times \mathbb{R}^{D^g_I} \to \mathbb{R}^1$ is a fully connected layer for attention scoring;
$n^{(k)}_I = W^e_I \phi^{(k)}_I$ and $n^{(b)}_I = W^e_I \phi^{(b)}_I$ are the graph node embeddings;
$W^e_I: \mathbb{R}^{L_I \times C} \to \mathbb{R}^{D^g_I}$ and $W^d_I: \mathbb{R}^{D^g_I} \to \mathbb{R}^{L_I \times C}$ are learnable linear transformations;
$D^g_I$ is the dimension of the graph node embeddings, and $\gamma \in [0, 1]$ is a decay rate.
Then, we attach the enhanced prompts to the source model and update the prompts and the graph networks based on Eq. \ref{eq:uom}

\subsection{Overall Process}
The overall process of the I-DiPT framework is summarized as follows.
For each incoming image, we first select uncertain and reliable tokens based on the uncertainty estimation mechanism. Then, we pre-initialize the image-specific and image-invariant prompts for the incoming image and enhance the prompts by the proposed PGD. 
The prompts and graph networks are updated jointly according to the Eq. \ref{eq:uom}.
After the adaptation, the updated prompts are inserted into the frozen source model to make final predictions.
The low-frequency component and the updated prompts are enqueued into the prompt bank following the FIFO principle.
The pseudo-code of I-DiPT is shown in Algorithm \ref{algorithm1}.

\begin{algorithm}[t!]
\caption{I-DiPT method.}
\small

\KwIn{Incoming test image $x_k$ from the fragment $\mathcal{R}_{ij}$, source model $f_\theta$, random initialized image-specific
and image-invariant prompts $\phi^{(k)}_S$, $\phi^{(k)}_I$, prompt bank $\mathcal B$,
graph networks $\mathcal G_S$, $\mathcal G_I$, hyperparameters $D$, $K\%$, $L_S$, $L_I$, $N_B$, $\beta$, $\gamma$, $D^g_S$, $D^g_I$.}
\KwOut{Adapted prediction $\hat y_a$.}

\begin{algorithmic}[1]

\STATE $\triangleright$ \textbf{Select uncertain and reliable image tokens:}
\STATE $U_j=\sqrt{\frac1D\sum_{d=1}^{D}\|p(e_{dj})-\mu_j\|^{2}}$
\STATE 
$\mathcal{T}_{U}\leftarrow$ indices of highest $K\%$ values in $\{U_j\}$  

$\mathcal{T}_{R}\leftarrow$ indices of lowest $K\%$ values in $\{U_j\}$

\STATE $\triangleright$ \textbf{Enhance the prompts with PGD:}
\IF{$k < N_B$} 
\STATE $\hat{\phi}^{(k)}_{S}$ = $\phi^{(k)}_S$, $\hat{\phi}^{(k)}_{I}$ = $\phi^{(k)}_I$
\ELSE
\STATE 
$\tilde{\phi}^{(k)}_{S}\!=\!\sum_{b=1}^{N_B}\omega_b\,\phi^{(b)}_{S}$
\STATE
Obtain $\hat{\phi}^{(k)}_{S}$ and $\hat{\phi}^{(k)}_{I}$ by Eq. \ref{eq:specific_graph_prompt} and Eq. \ref{eq:invariant_graph_prompt}
\ENDIF

\STATE $\triangleright$ \textbf{Update the prompts and graph networks:}
\STATE Obtain $y_{u}$ and $y_{r}$ using $f_{\hat{\phi}^{(k)}_{S}}$ and $f_{\hat{\phi}^{(k)}_{I}}$ based on $\mathcal{T}_{U}$ and $\mathcal{T}_{R}$
\STATE Obtain $\hat{y}_{w}$ using source model $f_\theta$ shown in Sec. \ref{sec:mased_learning}
\STATE Backward and update $\hat{\phi}^{(k)}_{S}, \hat{\phi}^{(k)}_{I}, \mathcal{G}_{S}, \mathcal{G}_{I}$ by Eq. \ref{eq:uom}

\STATE $\triangleright$ \textbf{Inference with updated prompts:}
\STATE $\hat{y}_{a}=f_{\hat{\phi}^{(k)}_{I},\hat{\phi}^{(k)}_{S}}\!\big(x_k\big)$

\STATE $\triangleright$ \textbf{Update the prompt bank:}
\STATE $k_{\text{new}}\!=\!\mathcal{A}_{L}(x_k),\;
       v_{\text{new}}\!=\!(\hat{\phi}^{(k)}_{S},\hat{\phi}^{(k)}_{I})$, 
       $\mathcal{B}.\text{enqueue}(k_{\text{new}},v_{\text{new}})$
\IF{$len(\mathcal{B})>N_{B}$}
  \STATE $\mathcal{B}.\text{dequeue}(k_\text{first},b_\text{first})$
\ENDIF
\RETURN $\hat{y}_{a}$
\end{algorithmic}

\label{algorithm1}
\end{algorithm}

\section{Experiments}
\label{sec:exp}
\subsection{Dataset and Experimental Setting}
\label{subsec:exp}
\subsubsection{Breast Cancer Histology Image Classification} 
\begin{figure}[!htbp]
    \centering
\includegraphics[width=\linewidth]{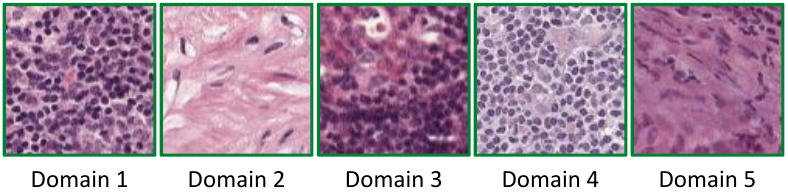}
    \caption{Examples from the five domains of Camelyon17 dataset.}
    \label{fig:came_example}
\end{figure}

We evaluate our method on the Camelyon17 dataset \cite{camlyon17}, which consists of over 450K histology image patches of size 96$\times$96 from five different medical sites, including: 1) 103,624 patches from Radboud University Medical Center (Domain 1); 2) 142,504 patches from Canisius-Wilhelmina Hospital (Domain 2); 3) 71,503 patches from University Medical Center Utrecht (Domain 3); 4) 36,416 patches from Rijnstate Hospital (Domain 4); and 5) 101,905 patches from Laboratorium Pathologie Oost-Nederland (Domain 5).
These samples exhibit heterogeneous appearances across the sites. The task of this dataset is to predict whether breast cancer is present in a histology image patch.
In our experiments, we take all samples from each site as a single domain. For each domain, we randomly split all samples to 60\% as the training set, 15\% as the validation set, and 25\% as the testing set at the patch level. 
The examples of Camelyon17 dataset are shown in Fig. \ref{fig:came_example}.

\subsubsection{Glaucoma Fundus Image Classification} 
\begin{figure}[!htbp]
    \centering
\includegraphics[width=0.9\linewidth]{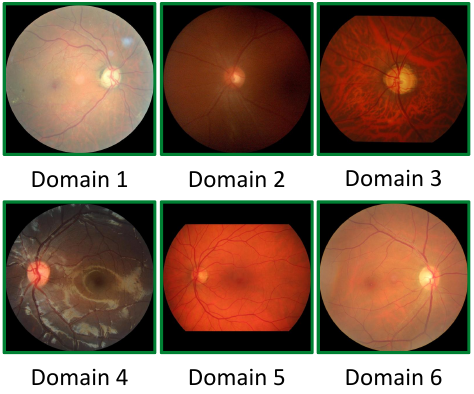}
    \caption{Examples of the fundus images from different domains.}
    \label{fig:glau_example}
\end{figure}

We also validate our method on the Standardized Multi-Channel Dataset for Glaucoma (SMCDG) \cite{glaudata}, which comprises over 12K fundus images from multiple medical sites. 
We follow a processing approach similar to previous work \cite{glauprocess}. 
Specifically, we select data from 11 available datasets in the SMCDG including: 1) 7,711 images from the EyePACS-AIROGS and OIA-ODIR datasets due to the severe class imbalance in the two datasets (Domain 1); 2) 139 images combined from CRFO-v4, JSIEC-1000, and LES-AV due to the similar intensity distribution and the limited scale of each dataset (Domain 2); 3) 101 images from the DRISHTI-GS dataset (Domain 3); 4) 400 images from the FIVES dataset (Domain 4); 5) 460 images from three datasets, DR-HAGIS, HRF, and PAPILA due to the same reason as Domain 2 (Domain 5); and 6) 401 images from the sjchoi86-HRF dataset (Domain 6).
All fundus images are preprocessed, including background cropping, centering, missing information padding, and resizing to 512$\times$512. 
Images from each domain are randomly split to the training (60\%), validation (15\%) and testing sets (25\%).
Finally, we constructed a 6-domain glaucoma classification dataset, with the task of determining whether a given fundus image is normal or indicative of glaucoma.
The examples of each domain are shown in Fig. \ref{fig:glau_example}.

\subsubsection{Experimental Setting} 
\label{sec:datacurate}
\begin{figure}[!htbp]
    \centering
\includegraphics[width=\linewidth]{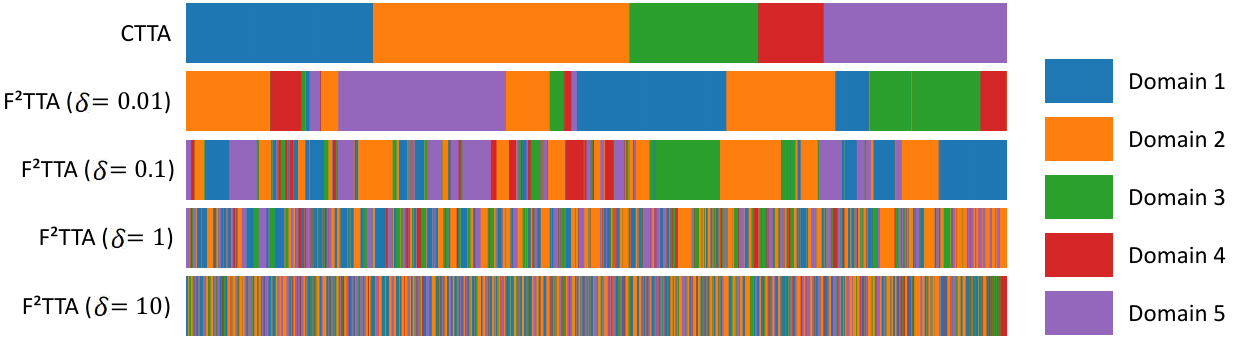}
    \caption{Visualization of data streams under CTTA and F\(^2\)TTA settings with different values of $\delta$.}
    \label{fig:datastream}
\end{figure}

In our F\(^2\)TTA scenario, we selected the training set of one domain as the source domain, while using the test sets of all other domains as target domains. 
For the test set of each domain, we randomly divided it into 100 domain fragments for the breast cancer classification task and 10 domain fragments for the glaucoma classification task.  
The length of each fragment follows Dirichlet distribution with parameter $\delta$.  
The fragments from different domains were interleaved randomly to form a test data stream, with distribution shifts occurring between consecutive fragments.  
Notably, the data stream also contained fragments from the test set of the source domain.  
We illustrate different data streams with varying $\delta$ values in Fig. \ref{fig:datastream}. 
We generated test data streams with $\delta=1$ to avoid extremely frequent domain changes that rarely occur in clinical practice. 
Eight data streams with $\delta=1$ were produced using different random seeds for subsequent experiments, except for the experiment in Sec. \ref{sec:data_stream_stable}.

\subsubsection{Evaluation Metric}
We adopted four commonly used evaluation metrics for classification tasks, including accuracy, precision, recall, and Area Under the Curve (AUC), to evaluate different methods.  
For all methods, the classification model was first updated, and the inference process was then repeated.  
We recorded the predictions of the model for each image during the adaptation process and calculated the evaluation metrics after traversing the entire data stream.  
The same evaluation protocol was employed for all competing methods to ensure fairness in comparisons.

\subsection{Implementation Detail}
Our method was implemented on a single NVIDIA GeForce 3090 GPU. We adopted ViT-B/16 \cite{vit} as our classification model. During the training of the model on source domain (i.e., source model), we used the Adam optimizer with learning rate $1e^{-4}$, epoch number 150, batch size 16 for glaucoma classification task, and batch size 512 for breast cancer classification task. 
During the adaptation process, we attached the image-specific and image-invariant prompts to each transformer layer of the ViT. 
We performed one-step adaptation for each batch of test data using the Adam optimizer with learning rate $1e^{-3}$ and batch size of 1 for both tasks.
We empirically set $L_{S}=8, L_{I}=4$ for both classification tasks. 
In UoM, we followed the MCDO setting of previous work \cite{cmae} and empirically set $D=10$ and $K\%=30\%$.
In PGD, we utilized a fully connected layer to implement the learnable linear transformation. 
We empirically set $N_{B}=20$ and $\beta=0.1$ for the prompt bank configuration, and $\gamma=0.9$ for the EMA strategy applied in the graph network $\mathcal{G}_I$.
We set the node dimensions of two graphs as $D^g_S=D^g_I=512$.

\subsection{Comparison with State-of-the-Arts}
\label{sec:main_result}
We conduct experiments to evaluate our method on breast cancer and glaucoma classification tasks. For both tasks, we select Domain 1 as the source domain and utilize other domains as the target domains. We re-implemented seven Test-Time Adaptation (TTA) methods for comparison, where all methods share the same network backbone (i.e., ViT-B/16).

\subsubsection{Compared Methods}
We build the baseline method: 1) \textbf{SourceOnly}, where the source model performs inference on target domains without adaptation. 
For Single-domain TTA (STTA) schemes, we evaluate: 1) \textbf{Tent} \cite{tent} updates the normalization layers by optimizing the entropy of model predictions; and 2) \textbf{DLTTA} \cite{dltta} dynamically adjusts the learning rate for each test image based on its style discrepancy. 
In addition, we conduct comparisons with five state-of-the-art Continual TTA methods: 1) \textbf{CoTTA} \cite{cotta} and \textbf{SAR} \cite{sar}, which enhance the supervision through test-time augmentation and noisy sample elimination, respectively; 2) \textbf{VPTTA} \cite{ctta-cvpr24} mitigates domain style discrepancies through visual prompts and normalization layer alignment; 3) \textbf{C-MAE} \cite{cmae} masks image patches with substantial distribution shift and reconstructs the histograms of oriented gradients (HOG) \cite{hog} features of masked patches; and 4) \textbf{DPCore} \cite{dpcore}, which considers dynamic domain changes caused by the unpredictability of weather conditions and adjusts visual prompts intelligently based on domain similarities.
Notably, DLTTA was integrated into the Tent framework.

\begin{table*}[!t]
    \centering
    \caption{The performance of comparison methods on the breast cancer classification task during adaptation on 8 different test data streams. Overall contains the average performance over the target domains and all domains. \textbf{Bold} font represents the best performance.}
    \renewcommand{\arraystretch}{1.5}
    \resizebox{\linewidth}{!}{
    \begin{tabular}{l|ccccc|c|c|c|c|c}
    \hline \hline
         &  \multicolumn{1}{c|}{Source} & \multicolumn{4}{c|}{Target}   &  \multicolumn{5}{c}{Overall}    \\
         \cline{2-11}
         & \multicolumn{1}{c|}{Domain 1}  & Domain 2  & Domain 3  & Domain 4  & Domain 5  & \multicolumn{1}{c|}{Target domains} & \multicolumn{4}{c}{All domains}  \\
    \hline \hline
    
    Metrics  & \multicolumn{5}{c|}{Accuracy (\%) $\uparrow$} & \multicolumn{1}{c|}{Accuracy (\%) $\uparrow$} & \multicolumn{1}{c|}{Accuracy (\%) $\uparrow$} & Precision (\%) $\uparrow$ & Recall (\%) $\uparrow$ & AUC (\%) $\uparrow$ \\
    \hline \hline

    SourceOnly & \multicolumn{1}{c|}{98.00 $\pm$ 0.00} & 64.64 $\pm$ 0.00 & 55.48 $\pm$ 0.00  & 77.96 $\pm$ 0.00  &  67.15 $\pm$ 0.00  & 66.31 $\pm$ 0.00 &  72.64 $\pm$ 0.00 & 84.43 $\pm$ 0.00  & 43.30 $\pm$ 0.00 & 86.98 $\pm$ 0.00 \\
    \hline

    Tent \cite{tent}  & \multicolumn{1}{c|}{41.14 $\pm$ 6.83 }  & 53.47 $\pm$ 0.75 & 44.78 $\pm$ 0.61  & 72.54 $\pm$ 1.25  & 59.08 $\pm$ 0.36  & 57.46 $\pm$ 0.38  & 54.20 $\pm$ 1.46  & 83.78 $\pm$ 6.06  & 3.62 $\pm$ 2.84 & 52.02 $\pm$ 5.99 \\

    DLTTA \cite{dltta} & \multicolumn{1}{c|}{53.31 $\pm$ 19.85} & 53.41 $\pm$ 1.83  & 48.04 $\pm$ 4.63 & 62.34 $\pm$ 13.16 & 56.49 $\pm$ 3.93 & 55.07 $\pm$ 3.39  & 54.72 $\pm$ 2.72 &  69.23 $\pm$ 19.17 & 31.74 $\pm$ 37.10 & 58.29 $\pm$ 4.36 \\
    \hline

    CoTTA \cite{cotta} & \multicolumn{1}{c|}{\textbf{98.19 $\pm$ 0.20}} & 65.85 $\pm$ 0.39  & 56.64 $\pm$ 0.76  & 78.36 $\pm$ 0.90  &  68.03 $\pm$ 0.50 & 67.22 $\pm$ 0.29 & 73.41 $\pm$ 0.22  & \textbf{84.96 $\pm$ 0.39}  & 44.85 $\pm$ 0.26 & 86.46 $\pm$ 0.19 \\

    SAR  \cite{sar} & \multicolumn{1}{c|}{97.89 $\pm$ 0.22} & 64.94 $\pm$ 0.36  & 55.22 $\pm$ 0.69 & 78.31 $\pm$ 0.93 & 67.13 $\pm$ 0.45  & 66.40 $\pm$ 0.29 & 72.70 $\pm$ 0.22 & 84.57 $\pm$ 0.45 & 43.27 $\pm$ 0.32 & 87.00 $\pm$ 0.16 \\

    VPTTA  \cite{ctta-cvpr24} & \multicolumn{1}{c|}{97.49 $\pm$ 0.21}  & 64.92 $\pm$ 0.34  & 55.78 $\pm$ 0.73  & 77.76 $\pm$ 1.09  & 67.54 $\pm$ 0.38  & 66.50 $\pm$ 0.34  & 72.70 $\pm$ 0.28  & 83.66 $\pm$ 0.44 & 43.72 $\pm$ 0.47 & 86.30 $\pm$ 0.16 \\

    C-MAE \cite{cmae} & \multicolumn{1}{c|}{97.15 $\pm$ 0.76} & 65.09 $\pm$ 3.07  & 53.75 $\pm$ 2.69 & 77.98 $\pm$ 1.30  & 67.08 $\pm$ 2.62  & 65.98 $\pm$ 1.83  & 72.21 $\pm$ 1.54  & 83.28 $\pm$ 1.05 & 42.86 $\pm$ 5.14 & 86.04 $\pm$ 0.96 \\

    DPCore \cite{dpcore} & \multicolumn{1}{c|}{92.41 $\pm$ 0.44} & 82.14 $\pm$ 0.30  & 67.83 $\pm$ 1.39 & 70.83 $\pm$ 1.43  & 80.30 $\pm$ 0.61  & 75.27 $\pm$ 0.49  & 78.70 $\pm$ 0.43  & 75.20 $\pm$ 0.43 & 74.68 $\pm$ 0.39 & 85.64 $\pm$ 0.35 \\
    \hline
    
    \textbf{I-DiPT (Ours)} & \multicolumn{1}{c|}{95.75 $\pm$ 0.60} & \textbf{84.90 $\pm$ 0.65} & \textbf{77.92 $\pm$ 1.17} & \textbf{78.89 $\pm$ 0.89} & \textbf{81.48 $\pm$ 0.28}  & \textbf{80.80 $\pm$ 0.46}  & \textbf{83.79 $\pm$ 0.38}  & 80.17 $\pm$ 1.06 & \textbf{80.27 $\pm$ 1.02} & \textbf{89.84 $\pm$ 0.25} \\
    \hline \hline
    \end{tabular}
    }
    \label{tab:main_exp_came17}
\end{table*}

\subsubsection{Results on Breast Cancer Histology Image Classification} 
We present the comprehensive evaluation on breast cancer classification in Table \ref{tab:main_exp_came17}. 
In SourceOnly, the source model demonstrates high performance on the source domain (Domain 1), but exhibits a significant performance decrease on the target domains, indicating a notable distribution shift between the target domains and the source domain. 
Predictably, STTA methods achieve worse performance than SourceOnly, even on the source domain. Although DLTTA accounts for the style discrepancies between each test sample and the source domain, it neglects the distribution shifts occurring between fragments from different domains, which interfere with the optimization objectives, thereby leading to error accumulation.
In contrast, CTTA methods achieve better performance than the STTA methods, mainly due to enhancing supervision or mitigating domain discrepancies to resist the distribution shifts between fragments. However, compared to SourceOnly, most CTTA methods still fail to adapt the model to the target domains in our F\(^2\)TTA setting, as evidenced by no performance improvement.
For instance, CoTTA and SAR rely on a stable data stream with periodically occurring shifts to acquire target-domain knowledge through the enhanced supervision.
Such stability cannot be achieved in our setting, leading to 0.77\% and 0.06\% overall accuracy improvement on all domains. 
VPTTA and C-MAE aim to learn domain-invariant knowledge from each test image where such knowledge is disrupted by the complex and unpredictable domain shifts, leading to limited performance.
DPCore adaptively selects prompt update strategies for each test sample based on its domain distance to the source domain and achieves a modest improvement compared to other CTTA methods.
However, DPCore lacks an explicit learning process for domain-invariant representations.
In contrast, I-DiPT effectively adapts the source model to each test image while learning domain-invariant representations to overcome the unpredictable shifts, achieving a significant improvement of 11.15\% in overall accuracy and 2.86\% in overall AUC on all domains over SourceOnly.

\begin{table*}[!t]
    \centering
    \caption{The performance of comparison methods on the glaucoma classification task during adaptation on 8 different test data streams. Overall denotes the average performance over the target domains or all domains, respectively. \textbf{Bold} font represents the best performance.}
    \renewcommand{\arraystretch}{1.5}
    \resizebox{\linewidth}{!}{
    \begin{tabular}{l|cccccc|c|c|c|c|c}
    \hline \hline
         &  \multicolumn{1}{c|}{Source} & \multicolumn{5}{c|}{Target}   &  \multicolumn{5}{c}{Overall}    \\
         \cline{2-12}
         & \multicolumn{1}{c|}{Domain 1}  & Domain 2  & Domain 3  & Domain 4  & Domain 5  & Domain 6  & \multicolumn{1}{c|}{Target domains} & \multicolumn{4}{c}{All domains}  \\
    \hline \hline
    
    Metrics  & \multicolumn{6}{c|}{Accuracy (\%) $\uparrow$} & \multicolumn{1}{c|}{Accuracy (\%) $\uparrow$} & \multicolumn{1}{c|}{Accuracy (\%) $\uparrow$} & Precision (\%) $\uparrow$ & Recall (\%) $\uparrow$ & AUC (\%) $\uparrow$ \\
    \hline \hline

    SourceOnly & \multicolumn{1}{c|}{92.10 $\pm$ 0.00} & 56.12 $\pm$ 0.00 & 68.32 $\pm$ 0.00 & 66.50 $\pm$ 0.00 & 59.35 $\pm$ 0.00 & 74.06 $\pm$ 0.00 & 64.87 $\pm$ 0.00 & 69.41 $\pm$ 0.00 & 59.03 $\pm$ 0.00 & 55.09 $\pm$ 0.00 &  71.10 $\pm$ 0.00 \\
    \hline

    Tent \cite{tent} & \multicolumn{1}{c|}{63.97 $\pm$ 4.11} & 58.09 $\pm$ 1.00 & 34.90 $\pm$ 5.04 & 62.72 $\pm$ 0.43 & 75.49 $\pm$ 0.40 & 74.97 $\pm$ 0.23 & 61.23 $\pm$ 0.95 & 61.69 $\pm$ 0.73 & 52.45 $\pm$ 14.07 & 5.50 $\pm$ 2.35 & 65.38 $\pm$ 1.45 \\

    DLTTA \cite{dltta} &  \multicolumn{1}{c|}{64.21 $\pm$ 4.83} & 57.82 $\pm$ 0.76 & 33.42 $\pm$ 2.79 & 62.66 $\pm$ 0.38 & \textbf{75.71 $\pm$ 0.22} & 75.0 $\pm$ 0.37 & 60.92 $\pm$ 0.66 & 61.47 $\pm$ 1.05 & 49.32 $\pm$ 13.99 & 5.11 $\pm$ 2.80 & 65.60 $\pm$ 1.49  \\
    \hline

    CoTTA \cite{cotta} & \multicolumn{1}{c|}{92.16 $\pm$ 1.45} & 53.69 $\pm$ 1.21 & 70.05 $\pm$ 1.96 & 66.00 $\pm$ 0.69 & 50.00 $\pm$ 3.02 & 72.51 $\pm$ 1.03 & 62.45 $\pm$ 0.83 & 67.40 $\pm$ 0.79 & 56.38 $\pm$ 0.64 & 62.22 $\pm$ 1.16 & 70.80 $\pm$ 0.27 \\

    SAR \cite{sar} & \multicolumn{1}{c|}{91.98 $\pm$ 2.17} & 56.83 $\pm$ 0.77 & 68.44 $\pm$ 0.63 & 65.59 $\pm$ 0.35 & 61.14 $\pm$ 0.80  & 74.31 $\pm$ 0.27 & 65.26 $\pm$ 0.27 & 69.72 $\pm$ 0.36 & 59.33 $\pm$ 0.40 & 53.85 $\pm$ 0.67 & 71.13 $\pm$ 0.16 \\

    VPTTA \cite{ctta-cvpr24} & \multicolumn{1}{c|}{\textbf{92.22 $\pm$ 2.05}} & 55.66 $\pm$ 1.27 & 68.19 $\pm$ 0.63 & 66.12 $\pm$ 0.38 & 59.97 $\pm$ 0.56 & 73.94 $\pm$ 0.35 & 64.78 $\pm$ 0.38 & 69.35 $\pm$ 0.39 & 58.77 $\pm$ 0.64 & 55.08 $\pm$ 0.52 & 71.15 $\pm$ 0.16 \\

    C-MAE \cite{cmae} & \multicolumn{1}{c|}{92.10 $\pm$ 2.09} & 56.29 $\pm$ 0.64 & 68.81 $\pm$ 0.75 & 65.66 $\pm$ 0.26 & 60.68 $\pm$ 2.66 & 74.47 $\pm$ 0.30 & 65.18 $\pm$ 0.64 & 69.67 $\pm$ 0.62 & 59.41 $\pm$ 0.88 & 54.05 $\pm$ 1.45  & 71.10 $\pm$ 0.18 \\

    DPCore \cite{dpcore} & \multicolumn{1}{c|}{88.74 $\pm$ 2.19} & 54.23 $\pm$ 1.96 & 67.45 $\pm$ 2.45 & 64.22 $\pm$ 1.79 & 56.49 $\pm$ 1.30 & 62.84 $\pm$ 1.03 & 61.04 $\pm$ 0.87 & 65.66 $\pm$ 0.76 & 56.01 $\pm$ 1.27 & 64.06 $\pm$ 1.16  & 69.87 $\pm$ 0.76 \\
    \hline
    
    \textbf{I-DiPT (Ours)} &  \multicolumn{1}{c|}{91.98 $\pm$ 1.76} & \textbf{64.03 $\pm$ 2.76} & \textbf{73.27 $\pm$ 1.05} & \textbf{67.00 $\pm$ 0.73} & 65.43 $\pm$ 1.17 & \textbf{77.31 $\pm$ 0.39} & \textbf{69.41 $\pm$ 0.73} & \textbf{73.17 $\pm$ 0.45} & \textbf{71.92 $\pm$ 2.84} & \textbf{64.83 $\pm$ 2.26} &  \textbf{72.49 $\pm$ 0.76} \\
    \hline \hline
    \end{tabular}
    }
    \label{tab:main_exp_glau}
\end{table*}

\subsubsection{Results on Glaucoma Fundus Image Classification}
In addition, we employ our method on fundus datasets and present the quantitative comparison results in Table \ref{tab:main_exp_glau}. 
This task is more challenging compared with the breast cancer classification task, as the test data includes images from 6 domains with more diverse and complex distribution shifts between the source domain and target domains. The model achieves only 69.41\% overall accuracy on all domains when directly applied to the test data.
The STTA methods exhibit extremely poor performance on the source and target domains, indicating their difficulty in handling the complex and unpredictable distribution shifts between domain fragments.
Although some CTTA methods achieve improvements over SourceOnly, their performance still lags behind our method due to the noisy and inefficient process of acquiring target domain knowledge from test data.
Our method achieves a 3.76\% higher overall accuracy and 1.39\% higher AUC compared to SourceOnly in such a challenging task.
The experimental results demonstrate the potential of our method on more target domains with more complex domain shifts.

\subsection{Detail Analysis}
\subsubsection{Effectiveness of Each Module}
We validate each module of our method through comprehensive experiments on the breast cancer classification task, including: 1) \textbf{Setting 1}, which utilizes a naive I-DiPT method without UoM or PGD; 2) \textbf{Setting 2}, which further incorporates UoM into I-DiPT; 3) \textbf{Setting 3}, which combines our I-DiPT with PGD; and 4) \textbf{Setting 4} representing the full implementation of I-DiPT.
Notably, we employ the pseudo labeling \cite{cotta} as the adaptation loss for the settings without incorporating UoM.
These results are shown in Tab. \ref{tab:ablation_module}. 
Compared to SourceOnly, Setting 2 shows an improvement in overall accuracy; however, Setting 1 leads to a decrease in overall accuracy. 
This result indicates that the image-level disentangled prompts learn to adapt the source model to incoming images under unpredictable shifts when guided by UoM. 
The performance of Setting 4 is further improved by PGD, which distills historical knowledge from previous prompts. 
Notably, compared to Setting 1, Setting 3 achieves negligible improvement, as the previous prompts fail to learn effective knowledge under the pseudo-labeling loss.

\begin{table}[!htbp]
    \centering
    \caption{Ablation study on breast cancer classification task to analyze the contribution of the proposed modules.}
    \resizebox{\linewidth}{!}{
    \begin{tabular}{l|c|c|ccc}
    \hline \hline
     \multirow{2}{*}{}  & \multirow{2}{*}{UoM}  & \multirow{2}{*}{PGD}   & \multicolumn{3}{c}{Accuracy (\%) $\uparrow$}  \\
     \cline{4-6}
     &  &    &  Source & Target & Overall   \\
    \hline
       SourceOnly & - & - & 98.00 & 66.31 & 72.64 \\
       Setting 1  &  - & - & 90.27 & 62.19 & 67.80 \\
       Setting 2  & \checkmark & - & 98.02 & 72.32 & 77.46 \\
       Setting 3  & - & \checkmark & 90.92 & 62.35 & 68.07 \\
       Setting 4 (I-DiPT)  & \checkmark & \checkmark & 95.75 & 80.80 & 83.79 \\
    \hline \hline
    \end{tabular}
    }
    \label{tab:ablation_module}
\end{table}

\subsubsection{Analysis of Uncertainty-oriented Masking}
\begin{figure}[!htbp]
    \centering
\includegraphics[width=\linewidth]{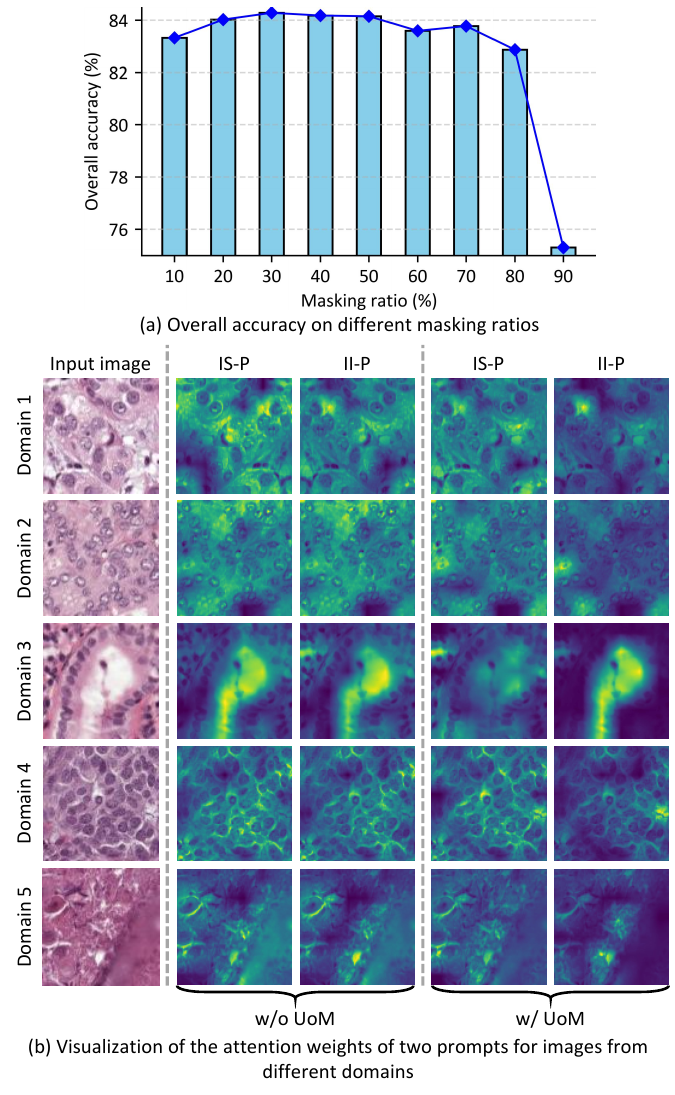}
    \caption{Analysis of Uncertainty-oriented Masking. (a) adaptation performance under different masking ratios ($K\%$); (b) extracted representations of the image-specific prompt (IS-P) and image-invariant prompt (II-P) with and without UoM.}
    \label{fig:uom_analysis}
\end{figure}

In our UoM, the masking ratio ($K\%$) is the most critical hyperparameter.
To investigate the sensitivity of UoM to the masking ratio, we compute the overall accuracy of all domains of I-DiPT on the validation set of the breast cancer dataset under $K\%$ ranging from 10\% to 90\% at intervals of 10\%. 
The experimental results are shown in Fig. \ref{fig:uom_analysis}(a).
We observe that the highest accuracy is achieved when the masking ratio is set to 30\%, and the accuracy fluctuation remains minimal within the range of 20\% to 70\%. 
This observation indicates that I-DiPT demonstrates robustness to the masking ratio when the masking ratio falls within an appropriate range.
As the masking ratio reaches extreme values (e.g., 90\%), the overall accuracy rapidly declines.
The reason is that the model struggles to make correct predictions based on few reserved patches.
Therefore, we set the masking ratio to 30\% in our method.

The UoM plays a crucial role in our I-DiPT as shown in Tab. \ref{tab:ablation_module}. To validate the knowledge representation capability of UoM, we obtain the attention maps of two prompts of the last MSA layer after the adaptation for the incoming image with and without UoM (i.e., pseudo-labeling loss), respectively.
The results are shown in Fig. \ref{fig:uom_analysis}(b).
When employing the pseudo-labeling loss, the two prompts fail to extract sufficient and discriminative knowledge from the input image, leading to highly similar representations.
In contrast, when combined with UoM, the image-specific prompts focus on high-level textures or appearance and the image-invariant prompts focus on the local image contents (e.g., the breast cancer cells or acini). 
These results indicate that, assisted by UoM, the prompts can effectively encode disentangled knowledge from a single test image.

\subsubsection{Analysis of Parallel Graph Distillation}
\begin{figure}[!htbp]
    \centering
    \includegraphics[width=0.8\linewidth]{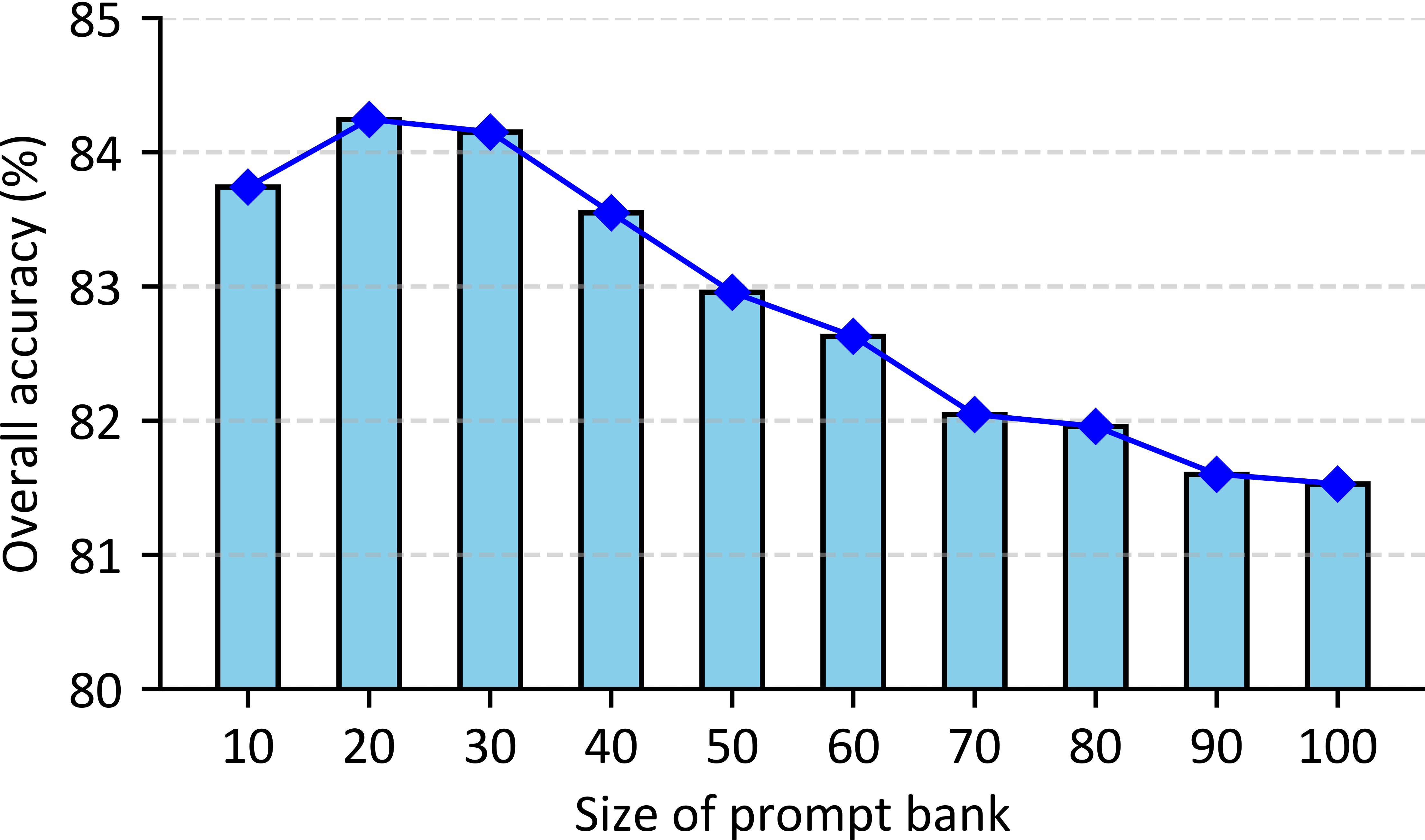}
    \caption{Effect of prompt bank size on the overall performance across all domains.}
    \label{fig:prompt_bank}
\end{figure}

We conduct experiments under different prompt bank sizes ranging from 10 to 100 at intervals of 10 and show the overall accuracy in Fig. \ref{fig:prompt_bank}. 
The accuracy reaches its peak at $N_B = 20$ and then decreases as $N_B$ gradually increases.
The results suggest that as the number of stored prompts increases, our PGD struggles to distill effective knowledge from historical prompts, resulting in performance degradation. 
Finally, we set $N_B = 20$.

We evaluate the role of the parallel graph networks ($\mathcal{G}_S, \mathcal{G}_I$) on the test set of breast cancer dataset.
We only use the pre-initialization term in Eq. \ref{eq:specific_graph_prompt}, without using the graph $\mathcal{G}_S$ for the image-specific prompt, and omit the graph $\mathcal{G}_I$ for the image-invariant prompt.
The experiment results are shown in Tab. \ref{tab:pgd_graph}.
Compared to SourceOnly, I-DiPT achieves improved accuracy when equipped with the pre-initialization mechanism for the image-specific prompts, as it aggregates knowledge from historical prompts. 
The performance is further enhanced by incorporating the graph networks, demonstrating their ability to capture the intricate non-linear relationships among the prompts.

\begin{table}[!htbp]
    \centering
    \caption{The performance of I-DiPT with and without the parallel graph networks.}
    \resizebox{0.8\linewidth}{!}{
    \begin{tabular}{c|ccc}
    \hline \hline 
       \multirow{2}{*}{Method}  & \multicolumn{3}{c}{Accuracy (\%) $\uparrow$}        \\
       \cline{2-4}
          & Source & Target & Overall   \\
       \hline \hline 
    SourceOnly  & 98.00 & 66.31 & 72.64   \\
    I-DiPT (w/o PGD)  & 98.02 &  72.32 & 77.46  \\
    I-DiPT (w/o $\mathcal{G}_S,\mathcal{G}_I$)  & 95.85 &  78.00 & 81.57  \\
    I-DiPT    & 95.75 & 80.80 & 83.79  \\
    \hline \hline 
    \end{tabular}}

    \label{tab:pgd_graph}
\end{table}

\subsubsection{Robustness to Different Source Domains}
\begin{figure}[!htbp]
    \centering
\includegraphics[width=\linewidth]{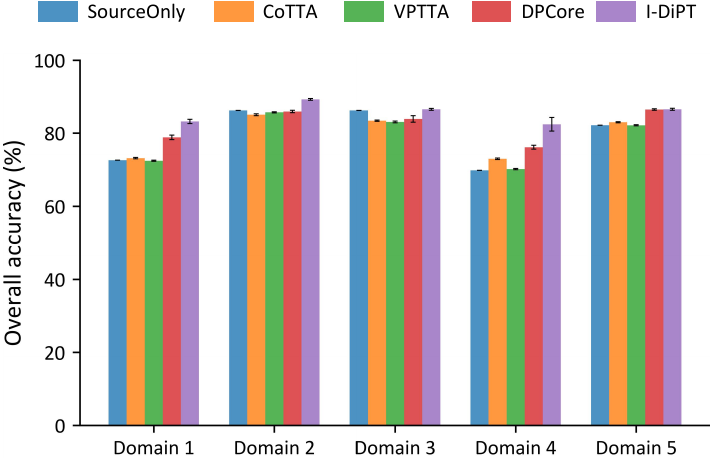}
    \caption{Performance comparison of different methods after training on different source domains on the breast cancer classification task.}
    \label{fig:domain_acc}
\end{figure}

Following the same data stream curation pipeline described in Sec. \ref{sec:datacurate}, we treat each domain as the source domain and generate eight different test data streams from the breast cancer dataset.
For each source domain, we train a source model and perform adaptation on the corresponding test data streams.
We report the experiment results in Fig. \ref{fig:domain_acc}.
Compared to three state-of-the-art CTTA methods, CoTTA, VPTTA and DPCore, I-DiPT achieves the highest overall accuracy for all the source domains.
I-DiPT demonstrates excellent robustness to various source domains, as it explicitly learns the domain-invariant representations.

\subsubsection{Computational Cost}
\label{sec:cost}
\begin{figure}[!htbp]
    \centering
\includegraphics[width=\linewidth]{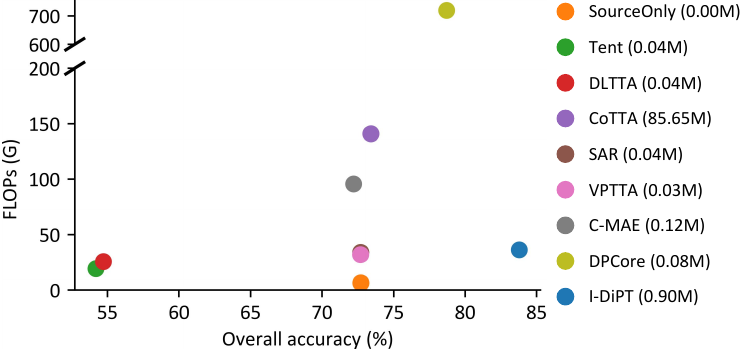}
    \caption{Computational cost of different methods in terms of FLOPs and learnable parameters (shown in parentheses) in relation to the overall accuracy on all domains of the histology dataset.}
    \label{fig:flops_acc}
\end{figure}

To evaluate the computational cost of I-DiPT, we calculate the floating point operations (FLOPs) and the number of learnable parameters during adaptation to a single image. 
We illustrate the computational cost and classification performance of different methods in Fig. \ref{fig:flops_acc}. 
Compared to other state-of-the-art TTA methods, I-DiPT achieves a significant improvement in the overall accuracy, with only a slight increase in computational cost primarily caused by the parallel graph networks.
The increase in computational cost is acceptable, as I-DiPT achieves the highest performance on all domains, especially in the challenging F\(^2\)TTA setting. 
I-DiPT achieves the best trade-off between the model performance and test-time computational cost.

\subsubsection{Analysis of Running Performance}
We evenly divide the entire test data streams of breast cancer data into 8 segments and calculate the average accuracy of different methods on each segment (i.e., running performance). Notably, we do not interrupt the adaptation process across different segments but merely record the accuracy. The results are shown in Tab. \ref{tab:run_acc}. 
During the adaptation process, the accuracy of I-DiPT steadily increases over time and achieves 88.88\% accuracy on the last segment, whereas other methods experience performance oscillations or declines. 
I-DiPT progressively accumulates domain-invariant knowledge from sequentially arriving test images, thereby steadily improving the source model performance on target domains.

\begin{table}[!htbp]
    \centering
    \caption{Performance of different methods over time across eight segments of test data streams.}
    \resizebox{\linewidth}{!}{
    \begin{tabular}{c|cccccccc}
    \hline \hline
        \multirow{2}{*}{Method} & \multicolumn{8}{c}{Accuracy per segment (\%) $\uparrow$} \\
    \cline{2-9}
         & 1 & 2 & 3 & 4 & 5 & 6 & 7 & 8 \\
    \hline \hline
       SourceOnly  & 73.22 & 71.00 & 75.72 & 71.48 & 73.53 & 74.12 & 78.02 & 64.40 \\
       CoTTA  & 73.87 & 71.95 & 76.14 & 72.26 & 74.06 & 74.06 & 77.71 & 66.73 \\
       VPTTA  & 72.64 & 70.77 & 75.60 & 71.79 & 73.44 & 73.50 & 78.16 & 64.40 \\
       DPCore  & 83.44 & 85.18 & 79.93 & 82.40 & 81.50 & 78.75 & 79.79 & 80.35 \\
       I-DiPT  & 80.61 & 80.18 & 84.17 & 86.89 & 85.09 & 85.49 & 87.56 & 88.88\\
    \hline \hline
    \end{tabular}
    }
    \label{tab:run_acc}
\end{table}

\subsubsection{Effect of Data Stream Stability}
\label{sec:data_stream_stable}
To evaluate the effect of data stream stability (i.e., the randomness in the length and arrival order of domain fragments), we generate eight test data streams from the validation set of the breast cancer dataset for each value of $\delta$, ranging from 0.01 to 10.
Figure \ref{fig:datastream} shows the visualization of data streams with different $\delta$ values.
In the F\(^2\)TTA setting, the domain shift between the adjacent fragments occurs more frequently and unpredictably as $\delta$ increases.
The results are presented in Tab. \ref{tab:different_delta}.
Compared to SourceOnly, I-DiPT shows a significant performance gain as $\delta$ increases, while other state-of-the-art CTTA methods, such as CoTTA and VPTTA, demonstrate little or no improvement even when faced with mild fragment changes $(\delta=0.01)$.
Notably, the accuracy of DPCore deteriorates rapidly in the most challenging case ($\delta=10$).
The experimental results indicate that I-DiPT exhibits excellent robustness to the unpredictable domain shifts across various instability levels.

\begin{table}[!htbp]
    \centering
    \caption{The overall accuracy of different methods on the test data streams under varying $\delta$ values.}
    \resizebox{0.8\linewidth}{!}{
    \begin{tabular}{c|cccc}
    \hline \hline 
       \multirow{2}{*}{Method}  & \multicolumn{4}{c}{Overall accuracy (\%) $\uparrow$}        \\
       \cline{2-5}
          & $\delta=0.01$ & $\delta=0.1$ & $\delta=1$ & $\delta=10$  \\
       \hline \hline 
    SourceOnly  & 72.44 & 72.44  & 72.44 & 72.44  \\
    CoTTA     & 73.34 & 73.60 & 73.21 &  73.22 \\
    VPTTA     & 72.60 & 72.29 & 72.50 & 72.55 \\
    DPCore    & 81.42 & 81.09 & 79.96 &  76.90  \\
    I-DiPT    & 84.53 & 84.53 & 84.24 & 83.62 \\
    \hline \hline 
    \end{tabular}}

    \label{tab:different_delta}
\end{table}

\section{Discussion}
\label{sec:discussion}
Data-driven deep neural networks demonstrate superior potential for medical image classification. Their success relies on collecting numerous images and corresponding annotations from multiple sites, which is highly impractical due to the stringent data sharing policies and intensive labor consumption. 
Therefore, when deploying a pre-trained model in real-world clinical environments, the model often suffers from severe performance degradation at test-time due to distribution shifts between the training datasets from limited sites (domains) and test data. 
Moreover, in clinical practice, data from multiple domains is often received in domain fragments without predefined lengths or arrival order, with domain shifts occurring between fragments.
To bridge this gap, this work studies a practical Free-Form Test-Time Adaptation (F$^{2}$TTA) paradigm that focuses adaptation on the free-form domain fragments. 
We propose Image-level Disentangled Prompt Tuning (I-DiPT) to adapt the source model to each test image while maintaining robustness against the unpredictable domain shifts.
To mitigate the limited data available for training image-level prompts, we enable the prompts to extract informative features from the incoming image and leverage historical knowledge preserved in previous prompts.

Existing STTA methods aim to adapt the model for each target domain under a static domain shift. This insight relies on accurate detection of the shift and is impractical because it is difficult to determine the domain identity of incoming data in advance \cite{pt4cl1}, especially in the F$^{2}$TTA setting. Directly applying STTA methods can lead to error accumulation due to domain shifts between target domains.
CTTA methods consider only a simplified setting of F$^{2}$TTA where data arrives sequentially in complete domain units. These methods designed for the simplified scenario exhibit limited adaptation capability as they cannot extract robust style or content representations due to the complex domain shifts in F$^{2}$TTA.
Our method employs image-level disentangled prompts to encode target domain knowledge, with the prompts enhanced by the proposed UoM and PGD. As shown in Section \ref{sec:main_result}, it delivers remarkable performance compared to existing TTA methods.

The trade-off between the computational cost and the adaptation performance is a critical issue. Intuitively, scaling test-time computation can bring performance improvement. 
However, the high computation overhead can hinder the application of the model in scenarios requiring real-time responses.
Moreover, updating the full parameters of the model on few or even a single image can cause the overfitting problem.
We design a parameter-efficient framework by updating only 0.9M parameters (about 1\% of the model's parameters) during the adaptation process, achieving the best performance on all domains with a slight increase in FLOPs compared to existing TTA methods. 
The computational cost is primarily incurred by the PGD, which plays an important role in knowledge retention. 
We propose the first method to address the challenges of F$^{2}$TTA without incurring excessive computational overhead.

This paper focuses on addressing distribution shifts caused by the acquisition differences from various domains, assuming no significant biological variation between these domains. 
However, in clinical practice, biological variation that can manifest as abnormal tissue or organ shapes often occurs across domains due to patient diseases \cite{d2ct}, leading to semantic shifts \cite{cgr}.
In the future, we will investigate introducing shape composition \cite{shapeprior} within I-DiPT to overcome this issue.
We also observe a mild performance degradation on the source domain after adapting the model to target domains, even though the test data stream contains images from source domain.
This is recognized as a trade-off between the model’s plasticity on incoming data and memorizability on previous data, which is a challenging problem studied by Continual Learning. 
Although our method achieves the best trade-off between the plasticity and memorizability compared to existing TTA methods, it is promising to further explore such trade-off under the complex and unpredictable domain shifts in F\(^2\)TTA.

\section{Conclusion}
\label{sec:conclusion}
We study a novel Free-Form Test-Time Adaptation (F\(^2\)TTA) task that focuses on adaptation to free-form domain fragments with domain shifts between the fragments. 
We propose the first F\(^2\)TTA framework, termed Image-level Disentangled Prompt Tuning (I-DiPT), which employs image-specific and image-invariant prompts to adapt the source model to incoming fragments and maintain robustness against the unpredictable shifts.
To mitigate the limited data available for updating the image-level prompts, we first propose Uncertainty-oriented Masking (UoM) to encourage the prompts to extract sufficient information from a single incoming image based on its spatial context relations. 
Then, we propose the Parallel Graph Distillation (PGD) that distills knowledge from historical prompts to further improve the adaptation performance.
Experiments indicate that our method outperforms other TTA methods in overcoming the unpredictable shifts.

\section*{Acknowledgments}
This work was supported by the Noncommunicable Chronic Diseases-National Science and Technology Major Project (2023ZD0501806), the National Key Research and Development Program (2024YFC2418201), and the Interdisciplinary Fund for Medical and Engineering of SJTU (YG2023LC08).

\bibliographystyle{model2-names.bst}\biboptions{authoryear}
\bibliography{refs}

\end{document}